\documentclass[10pt,twocolumn,letterpaper]{article}

\usepackage{iccv}
\usepackage{times}
\usepackage{epsfig}
\usepackage{graphicx}
\usepackage{amsmath}
\usepackage{amssymb}
\usepackage{bbm}
\usepackage{epsfig}
\usepackage{graphicx}
\usepackage{amsmath}
\usepackage{amssymb}
\usepackage{grffile}
\usepackage{algorithm,algcompatible,amsmath}
\usepackage{subfigure}
\usepackage{multirow}
\usepackage{booktabs}
\usepackage{enumitem}
\usepackage{mdwlist} 
\usepackage{fixltx2e}
\usepackage{lipsum}
\usepackage{bm}

\usepackage{color,xcolor}
\usepackage{dcolumn}

\usepackage[pagebackref=true,breaklinks=true,colorlinks,bookmarks=false]{hyperref}

\iccvfinalcopy 

\def\iccvPaperID{6660} 

\ificcvfinal\fi

\makeatletter
\def\thanks#1{\protected@xdef\@thanks{\@thanks
		\protect\footnotetext{#1}}}
\makeatother

\begin{document}

\title{ Parallel Rectangle Flip Attack: \\A Query-based Black-box Attack against Object Detection}

\author{Siyuan Liang\textsuperscript{1,2}, Baoyuan Wu\textsuperscript{3,4,$\dagger$}, Yanbo Fan\textsuperscript{5}, Xingxing Wei\textsuperscript{6}, Xiaochun Cao\textsuperscript{1,2,$\dagger$}\\
\textsuperscript{1}Institute of Information Engineering, Chinese Academy of Sciences, Beijing, China\\
\textsuperscript{2}University of Chinese Academy of Sciences, Beijing, China\\ 
\textsuperscript{3}School of Data Science, The Chinese University of Hong Kong, Shenzhen, China\\
\textsuperscript{4}Secure Computing Lab of Big Data, Shenzhen Research Institute of Big Data, Shenzhen, China\\
\textsuperscript{5}Tencent, Shenzhen, China \\ 
\textsuperscript{6}Institute of Artificial Intelligence, Hangzhou Innovation Institute, Beihang University, Beijing, China\\
{\tt\small\{liangsiyuan,caoxiaochun\}@iie.ac.cn;\tt\small wubaoyuan@cuhk.edu.cn;\tt\small fanyanbo0124@gmail.com;\tt\small xxwei@buaa.edu.cn}
}

\thanks{$\dagger$ indicates corresponding authors. Corresponds to {\tt wubaoyuan@cuhk.edu.cn} and {\tt caoxiaochun@iie.ac.cn}}

\maketitle
\ificcvfinal\thispagestyle{empty}\fi

\begin{abstract}
	Object detection has been widely used in many safety-critical tasks, such as autonomous driving. However, its vulnerability to adversarial examples has not been sufficiently studied, especially under the practical scenario of black-box attacks, where the attacker can only access the query feedback of predicted bounding-boxes and top-1 scores returned by the attacked model. 
	Compared with black-box attack to image classification, there are two main challenges in black-box attack to detection. 
	Firstly, even if one bounding-box is successfully attacked, another sub-optimal bounding-box may be detected near the attacked bounding-box. 
	Secondly, there are multiple bounding-boxes, leading to very high attack cost. 
	To address these challenges, we propose a Parallel Rectangle Flip Attack (PRFA) via random search. We explain the difference between our method with other attacks in Fig.~\ref{fig1}. Specifically, we generate perturbations in each rectangle patch to avoid sub-optimal detection near the attacked region. 
	Besides, utilizing the observation that adversarial perturbations mainly locate around objects' contours and critical points under white-box attacks, the search space of attacked rectangles is reduced to improve the attack efficiency. 
	Moreover, we develop a parallel mechanism of attacking multiple rectangles simultaneously to further accelerate the attack process. 
	Extensive experiments demonstrate that our method can effectively and efficiently attack various popular object detectors, including anchor-based and anchor-free, and generate transferable adversarial examples.

\end{abstract}

\section{Introduction}

Deep neural networks~\cite{szegedy2015going} has significantly boosted the developments of many important tasks, such as image classification~\cite{he2016deep, jia2020adv}, object detection~\cite{ren2016faster, redmon2016you, lin2017feature, lin2017focal, liu2016ssd}, 
medical image analysis \cite{chen2020effective}, \etc.  
For example, object detection has been successfully applied in many safety-critical scenarios, such as autonomous driving~\cite{levinson2011towards} and pedestrian detection~\cite{tian2015deep}, \etc. 
However, many studies~\cite{carlini2017adversarial, athalye2018obfuscated, zheng2019distributionally, carlini2017towards, Jia_2019_CVPR, bai2021targeted, survey, wei2018transferable, liang2020efficient, li2020toward, FanWLZLLY20, xu2019exact} have shown that the DNNs are vulnerable to adversarial attacks and may produce false predictions. 
If pedestrians or traffic signs are incorrectly detected in autonomous driving, it will cause substantial security risks in the real world. 

Compared with the massive works on attacking image classification, adversarial attacks against DNN-based object detection have not been thoroughly studied, especially in the black-box scenario, where only the predicted bounding-boxes and confidences of queries are accessible to the attacker. 
There are two main challenges in attacking the black-box object detection. 
Firstly, due to the widely used module called non-maximum suppression (NMS) in mainstream detectors, only the proposal with the highest confidence score is predicted, while other proposals with similar confidence in near locations are suppressed. Consequently, even if one predicted bounding box is successfully attacked (\ie, not detected), another sub-optimal bounding box may be detected in similar locations(as shown in Section D of the \textbf{Supplementary Material}).  
Secondly, the number of optimized targets (\ie, proposals) in object detection is much larger than that in classification~\cite{xie2017adversarial}. Take a $d$-dimensional image as an example, the computational complexity of the candidate proposals is $O(d^2)$, while the complexity of classification is $O(d)$. 
It will cause very high cost to attack object detector. 

To address above two challenges, we propose an effective and efficient query based black-box attack method against object detection, called \textbf{Parallel Rectangle Flip Attack (PRFA)}. 
Specially, we first search a rectangle patch randomly, and generate adversarial perturbations with the sign flipping along the vertical or horizontal direction, to present any detection in this attacked patch, including any sub-optimal proposals. 
Besides, we observe that adversarial perturbations generated by white-box attacks against detection with large magnitudes mainly locate at objects' contours and some critical points~\cite{wei2018transferable}. 
Inspired by this observation, the search space of attacked rectangles can be significantly reduced to improve the attack performance. 
Moreover, we design a parallel mechanism that multiple rectangles can be attacked simultaneously, which can further improve the attack efficiency.  
The proposed PRFA method achieves successful attack on many popular object detectors, including anchor-based (\eg, two-stage FR~\cite{ren2016faster} and one-stage YOLO~\cite{farhadi2018yolov3}), anchor-free model (\eg, FCOS~\cite{tian2019fcos}), and the ATSS model~\cite{zhang2020bridging}.

The main contributions of this work are threefold. 
\textbf{1)} To the best of our knowledge, this is the first work about query-based black-box attack against object detection. 
\textbf{2)} We propose an effective and efficient black-box attack method specially designed for attacking object detection, such that the main challenges including the sub-optical detection and high attack cost can be well addressed. 
\textbf{3)} Extensive experiments demonstrate the superior attack performance of our method on attacking many mainstream object detectors, including both anchor-based and anchor-free detectors.

\begin{figure}
	\begin{center}
		\includegraphics[width=1\linewidth]{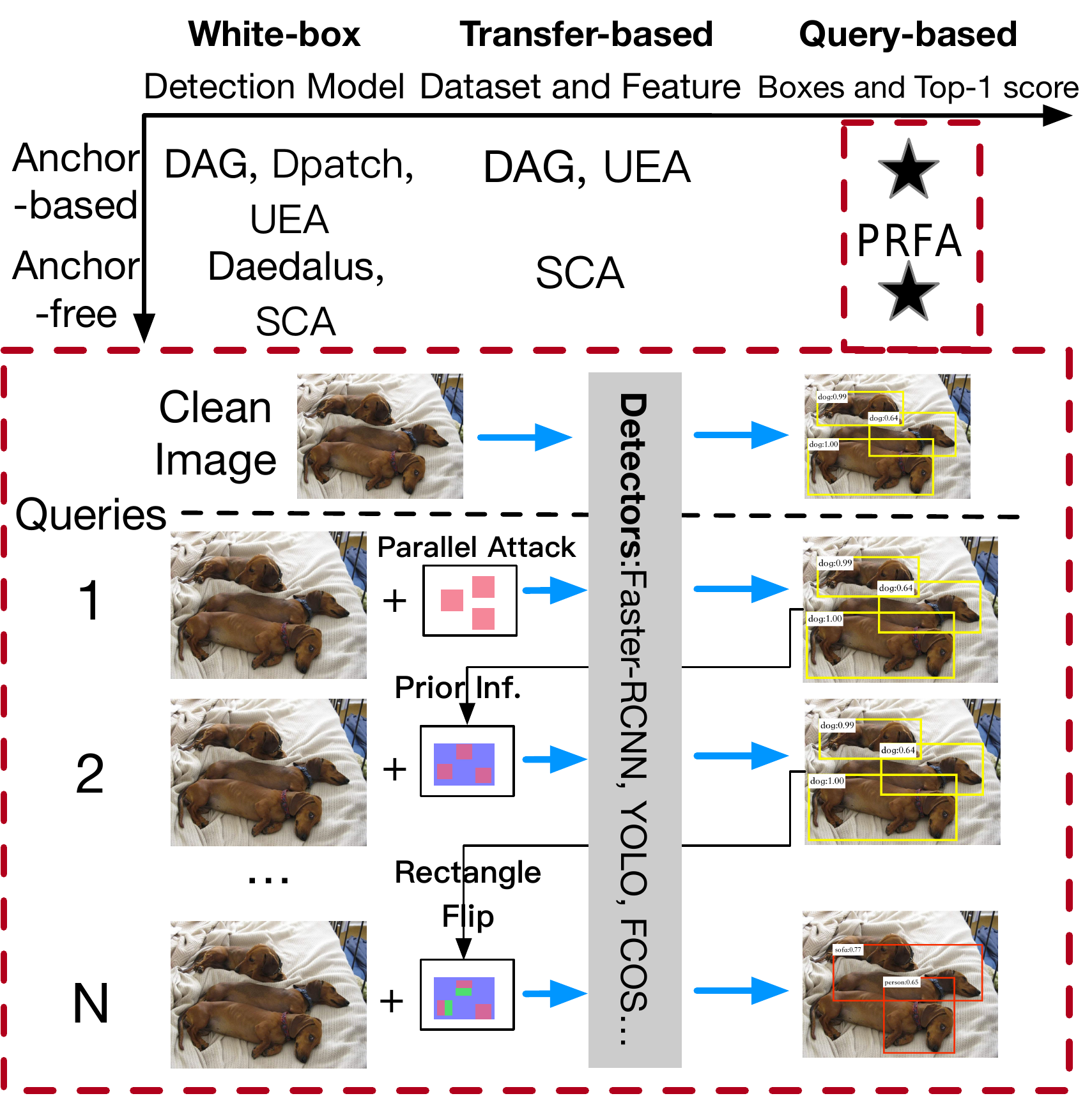}
	\end{center}
	\caption{We use the coordinate axis to show the taxonomy of the adversarial attack in object detection. Different from white box attacks and migration attacks, our method PRFA only relies on the prediction box and top-1 score output after NMS to attack through queries without gradients. PRFA can also attack anchor-free and anchor-based models at the same time.}
	\label{fig1}
	\vspace{-1.5em}
\end{figure}

\section{Related Work}
\subsection{Object Detection and White-box Attack}
Mainstream object detectors are mostly based on deep neural networks and can be roughly divided into two categories: anchor-based and anchor-free. The anchor-based detector divides the predefined sliding windows or proposals into positive or negative samples, then refines and classifies the prediction boxes. Due to the difference in the regression forms, it can be subdivided into the one-stage detector, such as SSD~\cite{liu2016ssd}, YOLOv2~\cite{redmon2017yolo9000} and two-stage detector, Faster-RCNN~\cite{ren2016faster}, Mask-RCNN~\cite{he2017mask}. The most representative anchor-free detector may be YOLOv1~\cite{redmon2016you}. YOLOv1 abandons the anchor and directly predicates the bounding box at the object's center. Since anchor-free detectors do not require extra parameter adjustment, these types of detectors have gained widespread popularity. Representative methods include CenterNet~\cite{duan2019centernet}, ExtremeNet~\cite{zhou2019bottom}, CornerNet~\cite{law2018cornernet} and FCOS~\cite{tian2019fcos}. Some methods focus on the gap between anchor-based and anchor-free detectors, such as ATSS~\cite{zhang2020bridging}, which improves the detection result by changing the sampling and IoU threshold calculation.
\\ \indent The existing adversarial attacks for object detection tasks are mainly white-box. As the first white-box attack method, DAG~\cite{xie2017adversarial} successfully fools Faster-RCNN by attacking the RPN network's proposals. \cite{bose2018adversarial} proposes a classification loss to train a GAN and generates adversarial perturbations for a face detector based on Faster-RCNN. ~\cite{li2018robust} designs a loss of predicting boxes and classification for the white-box attack. \cite{liu2018dpatch} successfully attacks the YOLO and Faster-RCNN models by generating an adversarial patch. Daedalus~\cite{wang2021daedalus} analyzes the vulnerability of NMS in the existing detection systems and attacks multiple detectors by generating many false-positive samples. \cite{liao2020category} proposes an attack method for the same category to attack the anchor-free detector. Since most detectors use the same feature extractor, UEA~\cite{wei2018transferable} generates transferable adversarial examples by destroying image features to attack Faster-RCNN and SSD models.
\\ \indent However, above adversarial attacks for object detection utilize the network's gradient more or less. In the real world, we cannot obtain the detector's gradient, which makes the attack very difficult. Therefore, we propose a query-based adversarial attack in a black-box scenario.

\subsection{Black-box Attack against Image Classification}
The black-box attack includes the transfer-based attack and the query-based attack. The attacker obtains adversarial examples by accessing the model’s outputs and modifying clean images. Using a transferable attack strategy, the adversary can train a substitute model~\cite{zhou2020dast} replacing the target model to get the adversarial gradient. DI-FGSM~\cite{xie2017adversarial} and TI-FGSM~\cite{dong2019evading} use `diverse inputs' or `translation-invariant' to generate more transferable adversarial examples against the defense models. Dispersion Reduction~\cite{lu2020enhancing} proposes an attack to minimize the `dispersion' of the feature map to enhance the transferability across different computer tasks. Other works focus on query feedback mechanisms against the logistics scores of all categories. ZOO~\cite{chen2017zoo} proposed a zero-order optimization method to estimate the gradient of the target model. Some studies are based on gradient's sign, such as~\cite{liu2018signsgd}, ZO-SignSGD extends SignSGD to the zero-order case and achieves black-box attacks. SignHunter~\cite{al2019sign} accelerates the convergence by combining the previous query results and converting the gradient estimation from continuous to binary. 
The Boundary \cite{brendel2018decision} and Evolutionary \cite{dong2019efficient} attack methods utilize the evolution strategy to gradually search the adversarial example that is close to the benign example in the scenario of decision-based black-box attack. 
The Sign Flip attack method \cite{chen2020boosting} proposes to search the perturbation by gradually shrinking the $\ell_\infty$-ball around the benign example and randomly flipping the signs of a few dimensions of the current perturbation. 
SquareAttack~\cite{andriushchenko2020square} based on a random search generates square-shaped perturbations at random positions. 
CG-ATTACK \cite{feng2020boosting} proposes to guide the search procedure by the conditional adversarial distribution, which is partially transferred from the distribution modeled by the c-Glow network \cite{lu2020structured} and trained on surrogate models. 

\indent Different from the classification, the optimization problem on object detection is complex. Suppose that $d$ denotes the number of pixels in one image, each proposal is determined by two pixels/coordinates, the complexity of object detection is $O(d^2)$ and of classification is $O(d)$. The detector’s outputs are prediction boxes after NMS and the top-1 score (probability of the top-1 label), making the black-box attack on object detection more like an intermediate setting between score-based and decision-based settings. How to achieve effective attacks with limited information and queries on detectors is our research focus.

\section{Parallel Rectangle Flip Attack}
This section will model a query-based black-box attack and introduce our method, which searches rectangle perturbations at random positions parallelly with fliping/reversing the sign of perturbations.
\subsection{Problem Formulation}
Suppose that a clean image $\bm{x}$ has $M$ recognition objects $\mathcal{O} = \{o_1, o_{2},...,o_{M}\}$. For each object $o_{m}, m=1,2,...,M$, is marked with ground-truth bounding box $g_{m}$ and a class label $y_{m} \in \{1,2, ...,Y\}$, where $Y$ is the number of classes. 
Object detection is an important computer vision task that predicts the position and a certain class(such as humans, transportation, or animals) of instances in digital images. An object detector $f(\bm{x}) \in \mathbb{R}^{N\times (4+1)}$ predicts the prediction boxes $b_{n},n=1,2,...,N$, and top-1 label $c_{n}, n=1,2,...,N$, with score $f_{C}$ (the probability after softmax normalization for predicted labels $C$) for $N$ objects. 
\\ \indent To generate an adversarial examples $\bm{\hat{x}}\in [0,1]^{d}$ which is regarded as adversarial examples with an $l_{p}$-norm of $\epsilon$ for the clean image $\bm{x}$, \emph{i.e.}, $||\bm{\hat{x}}-\bm{x}||_{p}\leq \epsilon$, and the goal is to make the IoU of all prediction boxes and ground-truth is less than a certain threshold or the labels of prediction boxes are classified incorrectly, that is, $\forall n\in N,\forall m\in M, (\mathrm{IoU}(b_{n}, g_{m}) < \mathrm{threshold}) \vee (c_{n} \neq y_{m}$). Here, IoU score is a standard performance measure for object detection, \emph{i.e.}, $\mathrm{IoU}(a,b)=(a \cap b)/(a \cup b)$, and the $\mathrm{threshold}$ is set to 0.5 for detection tasks. The task of find $\bm{\hat{x}}$ can be rephrased as solving the following optimization function:
\begin{flalign}
	\mathop{\arg\min}_{\bm{\hat{x}}\in [0,1]^{d}} ~ & H(f(\bm{\hat{x}}), B, Y) = \sum_{n=1}^N\sum_{m=1}^M[\mathrm{IoU}(b_{n}, g_{m}) \cdot \mathbbm{1}_{f_{c_{n}}\geq\zeta}
	\nonumber 
	\\
	& + \lambda\cdot (f_{c_{n}}-\max \limits_{C \neq y_m}f_C) \cdot \mathbbm{1}_{f_{c_{n}}<\zeta}], 
	\nonumber 
	\\ 
	s.t. ~ & ||\bm{\hat{x}}-\bm{x}||_{p}\leq \epsilon, f_{c_{n}}=f_{C}(\bm{\hat{x}}, b_{n}),  
	\label{eq1}
\end{flalign}
where the $\mathbbm{1}$ represents indicator function. The indicator function $\mathbbm{1}_{a} = 1$ if $a$ is true, otherwise 0. Since the attack satisfies one condition in Eq.~\eqref{eq1}, we can use the top-1 score $f_{c_{n}}$ as one of the judgment optimization formula one. Specifically, when the top-1 score is greater than the threshold $\zeta$, we consider reducing the IoU of the corresponding prediction box and ground-truth, and when it is less than the threshold $\zeta$, we optimize the top-1 score. $\lambda$ is a is a hyperparameter that adjusts balance. 
\\ \indent However, the optimization in Eq.~\eqref{eq1} needs to match all the prediction boxes with the ground-truth one by one, which makes the computation complexity reach $O(M*N)$ and costs a lot of time for one query. Therefore, we propose a category-based optimization function, which optimizes the prediction box and ground-truth under the same category, and sets the computational complexity to $O(M*N/|Y|)$. Because the detection dataset has too many categories, this will greatly improve one query speed. The $(N|y)$ means the index set of objects with the label $y$, \emph{i.e.}, $(N|y)=\{i|c_{i}=y, i=1,2,...,N\}$. The new optimization function $H$ is as follow:
\begin{flalign}
	\mathop{\arg\min}_{\bm{\hat{x}}\in [0,1]^{d}} ~ & \sum_{y=1}^{Y}\sum_{n=1}^{(N|y)}\sum_{m=1}^{(M|y)} \big[\mathrm{IoU}(b_{n}, g_{m}) \cdot \mathbbm{1}_{f_{c_{n}}\geq\zeta} + 
	\label{eq2}.
	\\
	& \lambda\cdot (f_{c_{n} = y}(\bm{\hat{x}},b_{n})-\max \limits_{C \neq y}f_C) \cdot \mathbbm{1}_{f_{c_{n}}<\zeta} \big],
	\nonumber 
	\\ 
	s.t. ~ & ||\bm{\hat{x}}-\bm{x}||_{p}\leq \epsilon, f_{c_{n}}=f_{C}(\bm{\hat{x}}, b_{n}).
	\nonumber
\end{flalign}
\begin{figure}
	\begin{center}
		\includegraphics[width=0.9\linewidth]{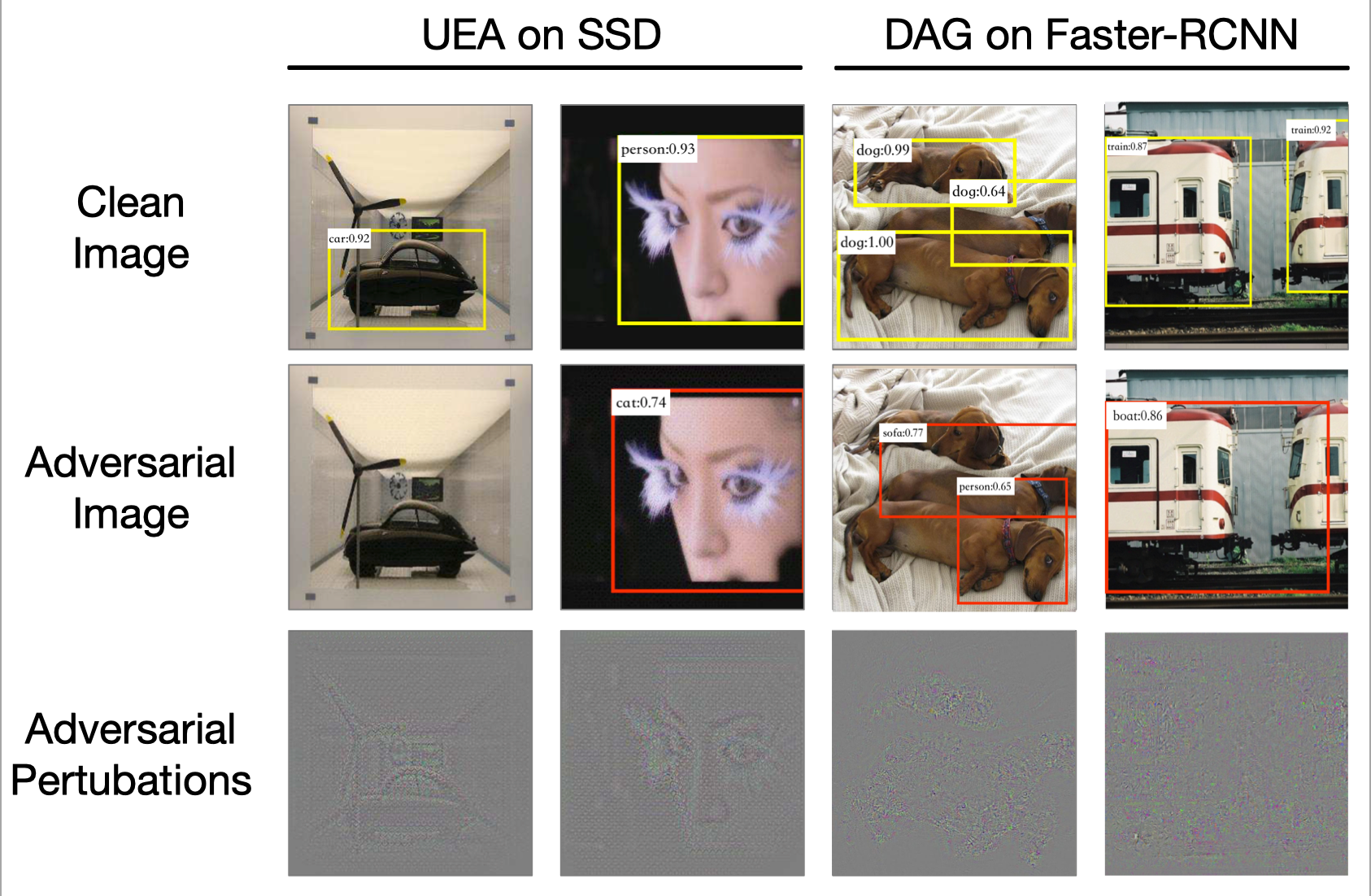}
	\end{center}
	\caption{The adversarial perturbations generated by the white-box attack methods UEA and DAG respectively on the SSD and Faster-RCNN. These perturbations are basically distributed at the contours and critical points of the object.}
	\label{fig2}
\end{figure}

\subsection{Prior-guided Dimensionality Reduction}\label{prior}

Dimensionality reduction has been shown to be effective to improve the efficiency of black-box attacking against image classification \cite{tu2019autozoom}. 
Inspired by this, we use the detector's prior information and the prior observation under the white-box attack to reduce the random search space. Although the anchor-based and anchor-free models are substantially different in anchor, they consider the objectness into prediction. Objectness is essentially a measure of the probability that an object exists in the region of interest. If the objectness is high, it means that the image window likely contains an object. We attack areas with high objectness instead of the entire image. Besides, we use DAG and UEA methods to observe the distribution of adversarial perturbations on different models. Although the attack methods and target models are different, the distributions of perturbations are concentrated on objects' critical areas or contours. As shown in Fig.~\ref{fig1}, we show the perturbations generated by DAG and UEA.
\\ \indent We use the prediction box or prior information to calculate an area with high objectness and perform a random search in this area. We optimize Eq.~\eqref{eq2} by generating rectangular perturbations through random sampling. The perturbations in this way are relatively close with white-box in position distribution. We considered three methods for calculating objectness, anchor-based priors(segmentation results from Mask-RCNN~\cite{he2017mask}), anchor-free priors(key points representation from RepPoints~\cite{yang2019reppoints}), and prediction boxes(outputs from detectors). Unlike the latter, the first two priors use the other detector's transferability to obtain critical areas. 

\subsection{Parallel Attack Accelerating Breadth Search}
We successfully modify the black-box attack methods such as SignHunter, SquareAttack, NES, and ZO-SignSGD to the object detection task by optimizing Eq.~\eqref{eq2}. Among them, the attack performance of SquareAttack is the most ideal. By observing the attack process of SquareAttack, we find that as the number of queries increases, the generated adversarial perturbations gradually gather around the object. This phenomenon shows that, under the constraints of Eq.~\eqref{eq2}, the black-box attack method can find the vulnerable pixels with a large number of queries. we can attack multiple positions parallelly in one query, thereby indirectly increasing the number of pixel searches, which can be a way to accelerate breadth search.
\\ \indent Next, we will theoretically analyze that it is not the best choice to generate adversarial perturbations $\mathbf{\delta}$ at only one random position at each query $q$. Suppose the optimized detector $f$ is the smoothness and has a Lipschitz gradient. There exists a constant $L$ satisfying:
\begin{equation}
	f(\bm{x}_{q+1}) - f(\bm{x}_{q})\leq\langle f^{'}(\bm{x}_{q}),\mathbf{\delta} \rangle+\frac{L}{2}||\mathbf{\delta}||^{2}\label{eq3}.
\end{equation}
According to the assumption 2 in~\cite{liu2021distributed}, the stochastic gradient is unbiased and with bounded variance, and its upper bound is a constant $\sigma^{2}$. 
The SquareAttack method satisfies the following:
\begin{equation}
	\frac{1}{Q}\sum_{q=0}^Q\mathbb{E}[||f^{'}(\bm{x}_{q})||^{2}] \lesssim \frac{f(\bm{x}_{0})-\mathbb{E}[f(\bm{x}_{Q+1})]}{Q\gamma} + \gamma L \sigma^{2}\label{eq4},
\end{equation}
where $\gamma$ is step size and when $\gamma=\frac{1}{L+\sigma\sqrt{QL}}$, the convergence rate is $O(1/\sqrt{Q})$. The $\lesssim$ means small and equal to up to a constant factor. The Eq.~\eqref{eq4} means that the number of iterations Q is large enough, and the random search algorithm will converge.
\begin{figure}
	\begin{center}
		\includegraphics[width=1\linewidth]{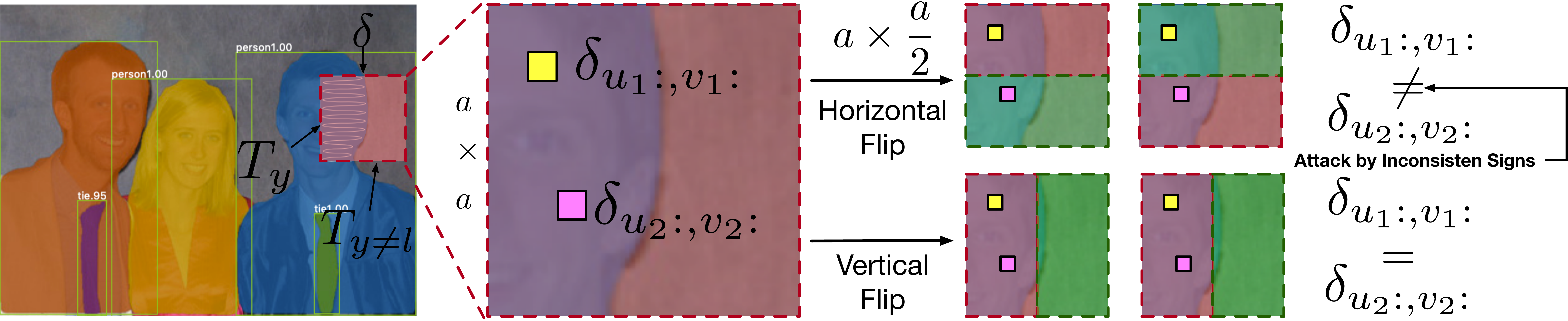}
	\end{center}
	\caption{The process of flipping perturbation's sign. We force the points with the same feature to be different by flipping sign, which will cause the detector to separate them to achieve an effective attack.}
	\label{fig3}
\end{figure}
\\ \indent We propose a parallel random search on the image. Each iteration $q$ can randomly sample $P$ positions in the search space $D$. In this way, we reduce the variance of the gradient estimation $\mathbb{E}[||g^{P}(\bm{x})-f^{'}(\bm{x})||^{2}] \leq (\frac{D-P}{D-1})\frac{\sigma^2}{P}$ and accelerate the convergence of the algorithm. At this time, the convergence can be expressed as:
\begin{equation}
	\frac{1}{Q}\sum_{q=0}^Q\mathbb{E}[||f^{'}(\bm{x}_{q})||^{2}] \lesssim \frac{L}{Q} + \frac{\sqrt{L}\sigma}{\sqrt{QP}}\label{eq5}.
\end{equation}
In Eq.~\eqref{eq5}, the influence of queries $Q$ on the algorithm convergence is still more significant than the number P of parallel attacks and the convergence rate is $O(1/\sqrt{QP})$ faster than Eq.~\eqref{eq4}. We adopted a parallel number scheduling strategy. Due to the small number of iterations in the early stage of the query, our sample number $P$ is the largest. As the number of iterations increases, we gradually decrease the value of $P$ until it is 1. See Section B in Supplementary Material for detailed proof.

\subsection{Rectangle Flip Attack for Depth Search}
Generally speaking, objects marked with a rectangular box in detection datasets have fixed sizes and scales. It is a typical prior for the detector to use predefined anchors with a fixed ratio. For example, Faster-RCNN uses 1:1, 1:2, and 2:1 anchor settings, and YOLO through k-means clustering learns different anchors from the training set. It is initially inspired by our experimental observation that given one region within one bounding box, perturbing different sub-regions with different noises is more likely to cause that this region is separately detected to different bounding boxes, compared to perturbing with similar noises. Furthermore, as shown in Fig.~\ref{fig3}, the square patch often covers one local region of one object, flipping sign horizontally or vertically may improve the perturbation diversity. Hence, this region is more likely to be detected falsely as different bounding boxes, causing the change of the original bounding box on this object.
\\ \indent Given an initial square-shaped $\mathbf{\delta} \in \mathbb{R}^{a\times a}$ with bound $\epsilon$ and a convolutional filter $w \in \mathbb{R}^{k\times k}$. Let $z=F((\bm{x}_{p} + \mathbf{\delta}) * w)$ denotes outputs of the CNN for updating $\mathbf{\delta}$, where $F$ denotes activation function and $\bm{x}_{p}$ is the clean patch. {$(m,n)$} represents coordinate in adversarial patch $\bm{\hat{x}}_{p}$. We can divided {$(m,n)$} into $|Y|+1$ cliques $\{T_i\}_{i=0}^{|Y|+1}$, and $(u,v)\in T_{i}$ denotes one point's coordinates in the $i$-th clique. The $\alpha$ is a constant greater than 0. The maximal change $l_{\infty}$-norm of $z$ represents:
\begin{equation}
	\begin{aligned}
		||z||_{\infty}=&\mathop{\max}\limits_{m,n}|z_{m,n}|\\=&\mathop{\max}\limits_{m,n}\vert F(\sum_{i,j=1}^k(\bm{x}_{p}+\mathbf{\delta})_{m-\left\lfloor \frac{s}{2} \right\rfloor+i,n-\left\lfloor \frac{s}{2} \right\rfloor+j} \cdot w_{i,j})\vert
		\\ \leq & \mathop{\max}\limits_{m,n}|F(\sum_{i,j=1}^k \mathbf{\delta}_{m-\left\lfloor \frac{s}{2} \right\rfloor+i,n-\left\lfloor \frac{s}{2} \right\rfloor+j} \cdot w_{i,j})+ \alpha \vert \label{eq6}.
	\end{aligned}
\end{equation}
\\ \indent For adversarial attack in object detection, the maximum change component of the correct label $y$ of an object contained in a patch should be smaller than the maximum change component of other classes. The optimization function $H$ in patch $\bm{\hat{x}_p}$ can be expressed as follows, the $\mathbf{\delta}_{u:,v:}$ is a shorthand in Eq.~\eqref{eq6}:
\begin{equation}
	\begin{aligned}
		\mathop{\min}_{\mathbf{\delta}} H = &\mathop{\min}_{\mathbf{\delta}} [\sum_{(u,v)\in T_{y}}^{T_{y}}\mathop{\max}|F(\sum_{i,j=1}^k \mathbf{\delta}_{u:,v:} \cdot w_{i,j})+ \alpha_{y}|\\ - & 
		\mathop{\max}\sum_{l \neq y}^{|Y|} [\sum_{(u,v)\in T_{l}}^{T_{l}}\mathop{\max}|F(\sum_{i,j=1}^k \mathbf{\delta}_{u:,v:} \cdot w_{i,j})+ \alpha_{l}|]]\label{eq7}.
	\end{aligned}
\end{equation}
Since Eq.~\eqref{eq7} is difficult to optimize, we combine adversarial perturbation and image semantic information to simplify Eq.~\eqref{eq7}. If two points locate in the same patch and belong to the same clique, then their perturbation should be the same, that is $(u_{1},v_{1})\in T_{l}$ and $(u_{2},v_{2})\in T_{l}$, then $\mathbf{\delta}_{u_{1}:,v_{1}:} = \mathbf{\delta}_{u_{2}:,v_{2}:}$. We propose an approximation to Eq.~\eqref{eq7}, as follows:
\begin{equation}
	\begin{aligned}
		\mathop{\min}_{\mathbf{\delta}} H \approx & \mathop{\min}_{\mathbf{\delta}} [\sum_{(u,v)\in T_{y}}^{T_{y}}\mathop{\max}|F(\sum_{i,j=1}^k \mathbf{\delta}_{y} \cdot w_{i,j})+ \alpha_{y}|\\ - & 
		\mathop{\max}\sum_{l \neq y}^{|Y|} [\sum_{(u,v)\in T_{l}}^{T_{l}}\mathop{\max}|F(\sum_{i,j=1}^k \mathbf{\delta}_{l} \cdot w_{i,j})+ \alpha_{l}|]]\label{eq8}.
	\end{aligned}
\end{equation}
The change value of the clique $T_y$ is maximum when the perturbations sign in every point is correct and consistent. Since the critical areas we attacked contain the object $y$, the clique labeled $T_y$ dominates. To minimize the Eq.~\eqref{eq8}, we can minimize the upper bound of the previous term. Points that belong to the same clique in a patch are highly similar in spatial location (close to value) and semantic feature (close in the property). Hence, if the sign of perturbations can change the semantic feature of one point, it will be effective for other points with high probability. Consequently, we can generate a rectangular perturbation by flipping the sign, which makes points in the same clique are inconsistent and pushes them into different classified classification boundaries. By this, we can minimize the Eq.~\eqref{eq8} for effective black-box.

\subsection{Generating Adversarial Examples}
Firstly, we use the detector's prior information in Sect.~\ref{prior} to calculate the high objectness area and determine the critical points to perform a random search. Then, we set the side length $a$ of the square perturbations, and the number of parallel points $P$ according to the dynamic scheduling algorithm. We first generate square-shaped perturbations with side length $a$ for each iteration. Next, we flip half of the square perturbations' sign (a rectangle with $a*a/2$ ) vertically or horizontally. We calculate scores by Eq.~\eqref{eq2} and update adversarial perturbations if the current score is greater than the optimal score. See Section A in Supplementary Material for the specific algorithm.

\begin{table*}[t]
	\caption{Ablation study on the Faster-RCNN model}
	\vspace{0.2cm}
	\label{tab1}
	\footnotesize
	\vspace{-4pt}
	\begin{center}
		\setlength{\tabcolsep}{1.7mm}{
			\begin{tabular*}{\textwidth}{ @{\extracolsep{\fill}} lcccccccc}
				\midrule
				\midrule
				\multirow{0}[4]{*}{Method} & $mAP$ &  $mAP_{50}$  &  $mAP_{75}$  &  $mAP_{S}$  & $mAP_{M}$  &  $mAP_{L}$  &  AQ \\
				\midrule
				\midrule
				Clean & 0.52 & 0.72 & 0.57 & 0.28 & 0.55 & 0.74 & $N/A$ \\
				Square Shaped & 0.28 & 0.50 & 0.27 & 0.21 & 0.38 & 0.32 & 3787 \\
				SS w. Prior & 0.26 & 0.48 & 0.25 & 0.20 & 0.30 & 0.33 & 3667 \\
				SS w. Prior \& Flip  & 0.24 & 0.42 & 0.23 & 0.18 & 0.27 & 0.28 & 3342\\
				SS w. Prior \& Parallel & 0.22 & \textbf{0.40} & 0.21 & 0.16 & 0.26 & 0.28 & 3513 \\
				\textbf{P}arallel \textbf{R}ectangle \textbf{F}lip \textbf{A}ttack & \textbf{0.21} & 0.41 & \textbf{0.19} & \textbf{0.16} & \textbf{0.26} & \textbf{0.27} & \textbf{3331}\\
				\midrule
				\midrule
		\end{tabular*}}
	\end{center}
	\vspace{-0.2cm}
\end{table*}

\begin{table*}[t]
	\vspace{-0.15em}
	\caption{Untargeted attack against object detectors. }
	\vspace{0.2cm}
	\label{tab2}
	\footnotesize
	\vspace{-4pt}
	\begin{center}
		\setlength{\tabcolsep}{1.7mm}{
			\begin{tabular*}{\textwidth}{ @{\extracolsep{\fill}} lcccccccccccccc}
				\midrule
				\midrule
				\multirow{2}[4]{*}{Method} &\multicolumn{7}{c}{Faster-RCNN~\cite{ren2016faster}}  & \multicolumn{7}{c}{ATSS~\cite{zhang2020bridging}} \\
				\cmidrule(lr){2-8} \cmidrule(lr){9-15} 
				& $mAP$ & $mAP_{50}$ & $mAP_{75}$ &  $mAP_{S}$  &  $mAP_{M}$  &  $mAP_{L}$  & $AQ$  & $mAP$ & $mAP_{50}$ & $mAP_{75}$ &  $mAP_{S}$  &  $mAP_{M}$  &  $mAP_{L}$  & $AQ$  \\
				\midrule
				\midrule
				Clean & 0.51 & 0.72 & 0.57 & 0.28 & 0.55 & 0.74 & $N/A$ & 0.54 & 0.73 & 0.60 & 0.32 & 0.58 & 0.74 & $N/A$ \\
				NES~\cite{ilyas2018black} & 0.49 & 0.69 & 0.54 & 0.26 & 0.52 & 0.69 & 4000 & 0.52 & 0.70 & 0.57 & 0.29 & 0.56 & 0.73 & 4000 \\
				ZSS~\cite{liu2018signsgd} & 0.49 & 0.71 & 0.54 & 0.20 & 0.52 & 0.71 & 4040 & 0.52 & 0.70 & 0.58 & 0.21 & 0.56 & 0.73 & 4040 \\
				SH~\cite{al2019sign} & 0.39 & 0.63 & 0.38 & 0.24 & 0.43 & 0.57 & 3987 & 0.40 & 0.55 & 0.44 & 0.20 & 0.40 & 0.59 & 3852 \\
				SA~\cite{andriushchenko2020square} & 0.28 & 0.50 & 0.26 & 0.21 & 0.38 & 0.32 & 3786 & 0.23 & 0.34 & 0.24 & 0.13 & 0.28 & 0.31 & 3505 \\
				PRFA & \textbf{0.21} & \textbf{0.42} & \textbf{0.19} & \textbf{0.16} & \textbf{0.26} & \textbf{0.27} & \textbf{3331} & \textbf{0.20} & \textbf{0.30} & \textbf{0.23} & \textbf{0.12} & \textbf{0.25} & \textbf{0.30} & \textbf{3500} \\
				\midrule
				\midrule
				\multirow{2}[4]{*}{Method} &\multicolumn{7}{c}{YOLOv3~\cite{farhadi2018yolov3}}  & \multicolumn{7}{c}{FCOS~\cite{tian2015deep}} \\
				\cmidrule(lr){2-8} \cmidrule(lr){9-15} 
				& $mAP$ & $mAP_{50}$ & $mAP_{75}$ &  $mAP_{S}$  &  $mAP_{M}$  &  $mAP_{L}$  & $AQ$  & $mAP$ & $mAP_{50}$ & $mAP_{75}$ &  $mAP_{S}$  &  $mAP_{M}$  &  $mAP_{L}$  & $AQ$  \\
				\midrule
				Clean & 0.45 & 0.70 & 0.47 & 0.16 & 0.47 & 0.65 & $N/A$ & 0.54 & 0.75 & 0.58 & 0.33 & 0.56 & 0.74 & $N/A$ \\
				NES~\cite{ilyas2018black} & 0.41 & 0.68 & 0.43 & 0.16 & 0.44 & 0.59 & 3958 & 0.53 & 0.73 & 0.57 & 0.23 & 0.56 & 0.77 & 4000 \\
				ZSS~\cite{liu2018signsgd} & 0.39 & 0.64 & 0.40 & 0.19 & 0.43 & 0.59 & 3958 & 0.52 & 0.71 & 0.56 & 0.27 & 0.57 & 0.74 & 4040 \\
				SH~\cite{al2019sign}  & 0.39 & 0.66 & 0.40 & 0.19 & 0.40 & 0.58 & 3911 & 0.27 & 0.40 & 0.31 & 0.09 & 0.37 & 0.64 & 3633 \\
				SA~\cite{andriushchenko2020square} & 0.25 & 0.49 & \textbf{0.22} & 0.15 & 0.31 & \textbf{0.34} & 3192 & \textbf{0.21} & 0.35 & \textbf{0.22} & \textbf{0.14} & \textbf{0.20} & \textbf{0.37} & 3578 \\
				PRFA & \textbf{0.24} & \textbf{0.46} & \textbf{0.22} & \textbf{0.13} & \textbf{0.29} & 0.36 & \textbf{2949} & 0.23 & \textbf{0.34} & 0.23 & 0.15 & 0.29 & 0.41 & \textbf{3395} \\
				\midrule
				\midrule

		\end{tabular*}}
	\end{center}
	\vspace{-1.8em}
\end{table*}

\begin{table*}[t]
	\vspace{-0.15em}
	\caption{Black-box transferability across different Detectors. The result with $^{*}num^{*}$ represents the adversarial examples from one black-box detector to attack itself, and the ordinary result represents transferability.}
	\vspace{0.2cm}
	\label{tab3}
	\footnotesize
	\vspace{-4pt}
	\begin{center}
		\setlength{\tabcolsep}{1.7mm}{
			\begin{tabular*}{\textwidth}{ @{\extracolsep{\fill}} lcccccccccccc}
				\midrule
				\midrule
				\multirow{2}[4]{*}{Adv. Dataset} &\multicolumn{6}{c}{Faster-RCNN}  & \multicolumn{6}{c}{ATSS} \\
				\cmidrule(lr){2-7} \cmidrule(lr){8-13} 
				& $mAP$ & $mAP_{50}$ & $mAP_{75}$ &  $mAP_{S}$  &  $mAP_{M}$  &  $mAP_{L}$  & $mAP$ & $mAP_{50}$ & $mAP_{75}$ &  $mAP_{S}$  &  $mAP_{M}$  &  $mAP_{L}$  \\
				\midrule
				\midrule
				Clean & 0.51 & 0.72 & 0.57 & 0.28 & 0.55 & 0.74 & 0.54 & 0.73 & 0.60 & 0.32 & 0.58 & 0.74 \\
				Faster-RCNN & $^{*}\textbf{0.21}^{*}$ & $^{*}\textbf{0.42}^{*}$ & $^{*}\textbf{0.19}^{*}$ & $^{*}0.16^{*}$ & $^{*}\textbf{0.26}^{*}$ & $^{*}\textbf{0.27}^{*}$ & 0.30 & 0.47 & 0.29 & \textbf{0.11} & 0.35 & 0.47 \\
				ATSS & 0.26 & \textbf{0.42} & 0.25 & \textbf{0.11} & 0.28 & 0.42 & $^{*}\textbf{0.20}^{*}$ & $^{*}\textbf{0.30}^{*}$ & $^{*}\textbf{0.23}^{*}$ & $^{*}0.12^{*}$ & $^{*}\textbf{0.25}^{*}$ & $^{*}\textbf{0.30}^{*}$ \\
				YOLO & 0.32 & 0.51 & 0.32 & 0.13 & 0.34 & 0.50 & 0.36 & 0.55 & 0.37 & 0.18 & 0.39 & 0.55 \\
				FCOS & 0.28 & \textbf{0.42} & 0.31 & 0.06 & 0.31 & 0.45 & 0.35 & 0.37 & 0.34 & 0.36 & 0.45 & 0.52 \\					
				\midrule
				\midrule
				\multirow{2}[4]{*}{Adv. Dataset} &\multicolumn{6}{c}{YOLO}  & \multicolumn{6}{c}{FCOS} \\
				\cmidrule(lr){2-7} \cmidrule(lr){8-13} 
				& $mAP$ & $mAP_{50}$ & $mAP_{75}$ &  $mAP_{S}$  &  $mAP_{M}$  &  $mAP_{L}$  & $mAP$ & $mAP_{50}$ & $mAP_{75}$ &  $mAP_{S}$  &  $mAP_{M}$  &  $mAP_{L}$  \\
				\midrule
				\midrule
				Clean & 0.45 & 0.70 & 0.47 & 0.16 & 0.47 & 0.65 & 0.54 & 0.75 & 0.58 & 0.33 & 0.56 & 0.74 \\
				Faster-RCNN & 0.22 & 0.39 & 0.22 & 0.08 & \textbf{0.23} & 0.41 & 0.27 & 0.44 & 0.26 & \textbf{0.09} & 0.31 & 0.46 \\
				ATSS & 0.24 & 0.43 & 0.24 & 0.10 & 0.26 & 0.41 & 0.28 & 0.45 & 0.28 & 0.11 & 0.33 & 0.47 \\
				YOLO & $^{*}0.24^{*}$ & $^{*}0.46^{*}$ & $^{*}0.22^{*}$ & $^{*}0.13^{*}$ & $^{*}0.29^{*}$ & $^{*}0.36^{*}$ & 0.35 & 0.54 & 0.36 & 0.11 & 0.38 & 0.54 \\
				FCOS & \textbf{0.16} & \textbf{0.35} & \textbf{0.08} & \textbf{0.07} & 0.28 & \textbf{0.32} & $^{*}\textbf{0.23}^{*}$ & $^{*}\textbf{0.35}^{*}$ & $^{*}\textbf{0.24}^{*}$ & $^{*}0.14^{*}$ & $^{*}\textbf{0.22}^{*}$ & $^{*}\textbf{0.36}^{*}$ \\	
				\midrule
				\midrule
		\end{tabular*}}
	\end{center}
	\vspace{-1.8em}
	\label{tab3}
\end{table*}

\section{Experiments}
\subsection{Experiments Settings}\label{exper setting}
We will introduce the experiment settings from four aspects: detection datasets, evaluation criteria, targeted models and parameters setting.
\\ \textbf{Detection Datasets} The MS-COCO~\cite{lin2014microsoft} is the most challenging object detection datasets today. A large number of object detectors use MS-COCO as a benchmark to evaluate the model's performance. MS-COCO includes 118k images for the training set and 5k images for the validation set. The objects in these pictures are divided into 80 categories. In order to ensure fairness, we attack the validation set. Specifically, we selected 
first 100 images as clean images for the black-box attack according to the MS-COCO API's loading order. These samples include a wealth of object instances, such as small objects or dense objects.
\\ \textbf{Evaluation Criteria} AP is defined as the average detection accuracy under different recall rates, and we usually evaluate it in a category-specific way (the mean AP, \textbf{mAP}). Refer to the detector, we use the evaluation criteria provided by MS-COCO, for example, $mAP$( averaged over multiple IoU
threshold between 0.5 and 0.95), $mAP_{50}$(mean AP at IoU=0.5), $mAP_{75}$(mean AP at IoU=0.75), $mAP_{S}$($area<32^{2}$), $mAP_{M}$($32^{2}<area<96^{2}$), $mAP_{L}$($area>96^{2}$). An effective black-box attack means a small $mAP$. In terms of algorithm efficiency, we use \textbf{AQ(average queries)} to evaluate the convergence of the algorithm. We hope to minimize the number of queries.
\\ \textbf{Targeted Models} We selected four representative detectors as targeted models. The first type is anchor-based. We chose the two-stage detector Faster-RCNN with ResNet50 and the YOLOv3 model with DarkNet53 as backbones. The second type belongs to anchor-free. We chose the FCOS model with ResNet50 as the backbone. We also selected ATSS, a detector that can adaptively select positive and negative samples. This model can eliminate the performance difference between anchor-based and anchor-free algorithms. The above models and codes are based on the open-source mmdetection library.
\\ \textbf{Parameters Setting} Given images of size $w*h$, the length $a$ of square is $\sqrt{e*w*h}$, $e \in [0, 1]$. The $e$ is set to 0.05, and we halve it at query $q \in \{20, 100, 400, 1000, 2000, 4000, 8000\}$. The parallel $P$ is 4 in the initial stage, and we halve them at $q \in \{20, 100, 1000, 2000\}$. In Eq~\eqref{eq2}, the $\zeta$  is 0.90 and the $\epsilon$ is 0.05. The $\mathrm{threshold}$ for $\mathrm{IoU}$ is 0.50. The  prior information from the same type of detector is more beneficial to attack similar detectors, our PRFA in all reported experiments only utilized the attacked model's outputs.

\begin{figure}
	\begin{center}
		\includegraphics[width=1.0\linewidth]{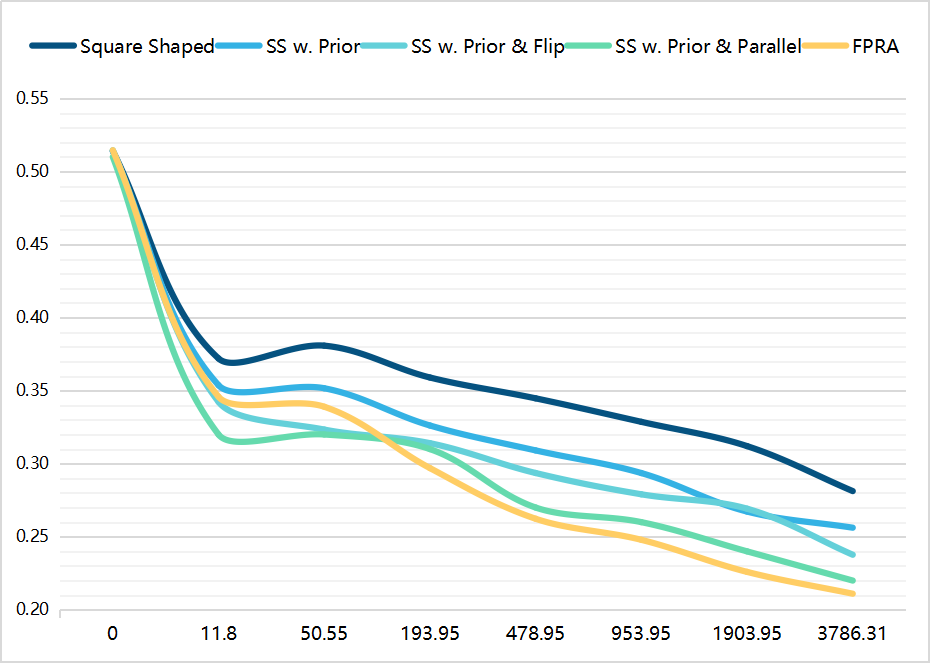}
	\end{center}
	\caption{$mAP$ changes $w.r.t.$ the number of queries for different attacks. FPRA achieves the fastest convergence and the most effective attack.}
	\label{fig4}
\end{figure}

\subsection{Abaltion Study}
We discuss the effectiveness of each component of the PRFA method through ablation experiments. In Tab.~\ref{tab1}, we show the ablation experiment on the Faster-RCNN model. `Clean' represents the detection result of Faster-RCNN on clean images. `Square Shaped' means that we only use square perturbations to perform random searches across the entire picture. By adding prior information, we can limit the space of random search. We add the prior restriction `SS w. Prior' to SS (short for square-shaped) to reduce the search space. Next, we generate rectangular perturbations by flipping the perturbations' sign. The method `SS w. Prior \& Flip' can be regarded as a combination of depth search and breadth search. Finally, we combine a parallel search attack to accelerate the algorithm's convergence and provide a parallel rectangle flip attack algorithm, namely PRFA.
\\ \indent The contribution of `Flip' is evaluated by attack performance (\eg, mAP reduction) and queries. 
As shown in Fig.~\ref{fig4}, in terms of mAP, the values of `SS w. Prior', `SS w. Prior \& Flip' and PRFA are 0.26, 0.24 and 0.21, respectively; in terms of queries, the values of these three methods are 3666, 3342 and 3331, respectively. `Flip' contributes 29\% to the mAP reduction and 97\% to the query reduction. `Flip' can also accelerate the convergence compared to `SS w. Prior' Therefore, the attack problem can be regarded as a trade-off between breadth search and depth search. The Fig.~\ref{fig4} shows the change in $mAP$ of each method over the number of iterations. By introducing parallel attacks, we effectively accelerated the model's convergence (yellow line) and got the best attack results.
\begin{figure}
	\begin{center}
		\includegraphics[width=1.0\linewidth]{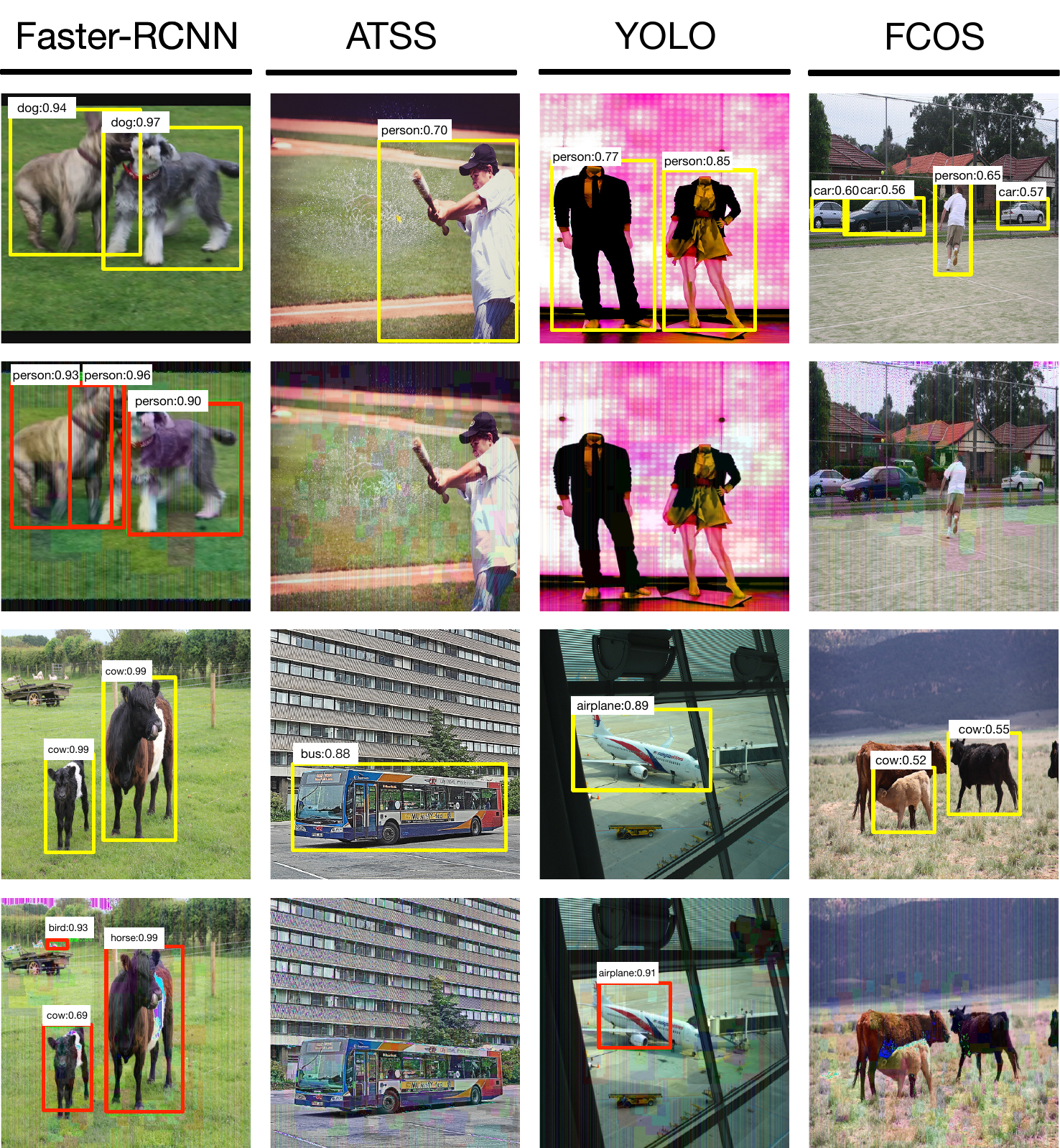}
	\end{center}
	\caption{The detection results of the four detectors on clean images (yellow lines) and adversarial samples (red lines).}
	\label{fig5}
\end{figure}

\subsection{Untargeted Attacks against Detectors}
We evaluated four state-of-the-art black-box attack algorithms and our method PRFA on the four detectors according to the settings in Sect.~\ref{exper setting}. The four methods are the NES black box attack combining PGD and NES strategies, the ZO-SignSGD(ZSS) algorithm combining zero-order optimization and sign stochastic gradient descent, the SignHunter(SH) method that converts the gradient estimation from a continuous problem to a binary problem, and the SquareAttack(SA) that generates square perturbations at a random location. We limited the number of queries for these methods, where NES is 4000, ZO-SignSGD is 4040, and SignHunter, SquareAttack, and PRFA are all 4000.
\\ \indent The performance of NES is the worst, and it can hardly affect the existing object detector. The performance of ZO-SignSGD on the models Faster-RCNN, ATSS, and FCOS is also inferior. He can only attack the YOLOv3 model. Although SignHunter can attack the existing object detector to a certain extent, we found that it uses nearly 4000 queries, which could not effectively converge with our query limits. SquareAttack and our method showed the effectiveness of the attack on all four models. Nevertheless, our method has a better attack effect on Faster-RCNN and ATSS, especially on Faster-RCNN, which drops 7 points and fewer queries (about 440). In the same attack effect (YOLOv3 and FCOS), the number of queries of our method is also better than the SquareAttack method.

\indent In summary, our method can achieve fewer queries or more effective attack, which means that our method will have faster convergence and search for more effective perturbations.

\subsection{Black-box Transferability across Models}
Finally, we investigate black-box transfer, \emph{i.e.}, using the perturbations generated by a black-box detector to attack other detectors. In Tab.~\ref{tab3}, the result with $^{*}num^{*}$ represents the adversarial examples from one black-box detector to attack itself, and the ordinary result represents transferability. 
\\ \indent In Tab.~\ref{tab3}, The attack effects of the three models Faster-RCNN, ATSS, and FCOS are robust. The reason is that they can resist attacks from other datasets. Among them, the most robust method is ATSS, which achieves an effect of at least 0.30 mAP on other datasets. The indicators of Faster-RCNN on other datasets are slightly lower than FCOS by 1-2 percentage points. Among all the models, YOLO is the most accessible model to attack. The adversarial examples generated by other models can attack him effectively, even surpassing the adversarial examples generated by itself.
\\ \indent Here we compare with three mentioned transfer methods, including `Dispersion', `DIFGSM' and `TIFGSM'. They firstly conduct white-box attacks on Faster-RCNN. 
Specifically, `Dispersion' attacks intermediate features of the ResNet50 backbone, while `DIFGSM' and `TIFGSM' attack the bounding box loss. Then, the obtained perturbations are transferred to attack the black-box target models ATSS, YOLOv3 ,and FCOS, respectively. The mAP values of these three methods and our PRFA method are $(0.05, 0.01, 0.02, 0.21)$ (PRFA conducts black-box attack) for Faster-RCNN, $(0.51, 0.48, 0.39, 0.3)$ for ATSS, $(0.43, 0.43, 0.4, 0.22)$ for YOLOv3, $(0.51, 0.48, 0.42, 0.27)$ for FCOS, respectively. 
PRFA significantly surpasses these transfer-based black-box methods.

\section{Conclusion}
In this paper, we propose a query-based black-box attack with the prediction boxes and top-1 scores. We adapt the existing black-box attack method to detection as the baseline. We propose a parallel rectangle flip attack via random search. We use prior information from detectors to reduce the search space by observing the white-box perturbations' distribution. We regard the optimization problem of searching perturbations for the detector as a trade-off between breadth search and depth search. We accelerate the model's convergence by parallel random walks in the search space in terms of breadth. We obtain a better attack effect by flipping the perturbations' sign locally to generate rectangular perturbations in terms of depth. Experiments show that our proposed method PRFA can attack mainstream object detectors and generate transferable adversarial examples.

\section*{Acknowledgement}
Supported by the National Key R\&D Program of China under Grant 2019QY(Y)0207, National Natural Science Foundation of China (No.62025604, 61861166002, U1936208), Open Project Program of State Key Laboratory of Virtual Reality Technology and Systems, Beihang University (No.VRLAB2021C06). Baoyuan Wu is supported by the Natural Science Foundation of China under grant No.62076213, the university development fund of the Chinese University of Hong Kong, Shenzhen under grant No.01001810, the special project fund of Shenzhen Research Institute of Big Data under grant No.T00120210003, and 2021 Tencent AI Lab Rhino-Bird Focused Research Program. Xingxing Wei is supported by National Natural Science Foundation of China (No.62076018, 61806109) and CCF-Tencent Open Research Fund.

{\small
	\bibliographystyle{ieee_fullname}
	\bibliography{egbib}
}

\end{document}